\documentclass[journal]{IEEEtran}
\usepackage[T1]{fontenc}
\usepackage{graphicx}
\usepackage{url}
\usepackage{amsmath,amsfonts,verbatimbox}
\usepackage{cite}
\usepackage{color}
\usepackage{subfig}
\usepackage{multirow}
\usepackage[ruled,norelsize,linesnumbered]{algorithm2e}

\IEEEoverridecommandlockouts

\makeatletter
\newcommand{\removelatexerror}{\let\@latex@error\@gobble}
\g@addto@macro{\@algocf@init}{\SetKwInOut{Parameter}{Parameters}}
\makeatother

\newcommand{\ie}{{\it i.e.}}

\newcommand{\reals}{\mathbb{R}}
\newcommand{\Expec}{\mathop{\mathbb{E}{}}}

\newcommand{\ecal}{\mathcal{E}}
\newcommand{\lcal}{\mathcal{L}}
\newcommand{\ncal}{\mathcal{N}}
\newcommand{\scal}{\mathcal{S}}
\newcommand{\tcal}{\mathcal{T}}
\newcommand{\ucal}{\mathcal{U}}
\newcommand{\vcal}{\mathcal{V}}

\newcommand{\argmax}{\mathop{\rm argmax}}

\newcommand{\bt}[1]{\textcolor{black}{#1}}
\newcommand{\bbt}[1]{\textcolor{black}{#1}}

\title{Multi-Agent Deep Reinforcement Learning for Large-scale Traffic Signal Control}
\author{
Tianshu Chu, Jie Wang, Lara Codec\`a, and Zhaojian Li~\IEEEmembership{Member,~IEEE}
\thanks{\quad Tianshu Chu and Jie Wang are with the Department of Civil and Environmental Engineering, Stanford University, CA 94305, USA. Lara Codec\`a is with the Communication Systems Department, EURECOM, Sophia-Antipolis 06904, France. Her work was partially funded by the ANR project ref. No.~ANR-11-LABX-0031-01 and EURECOM partners: BMW Group, IABG, Monaco Telecom, Orange, SAP, ST Microelectronics and Symantec. Zhaojian Li is with the Department of Mechanical Engineering, Michigan State University, East Lansing, MI 48824, USA.
Email: {\tt cts198859@hotmail.com}, {\tt jiewang@stanford.edu}, {\tt Codeca@eurecom.fr}, {\tt lizhaoj1@egr.msu.edu}.}%
}

\begin{document}
\maketitle

\begin{abstract}
Reinforcement learning (RL) is a promising data-driven approach for adaptive traffic signal control (ATSC) in complex urban traffic networks, and deep neural networks further enhance its learning power. However, centralized RL is infeasible for large-scale ATSC due to the extremely high dimension of the joint action space. Multi-agent RL (MARL) overcomes the scalability issue by distributing the global control to each local RL agent, but it introduces new challenges: now the environment becomes partially observable from the viewpoint of each local agent due to limited communication among agents. Most existing studies in MARL focus on designing efficient communication and coordination among traditional Q-learning agents. This paper presents, for the first time, a fully scalable and decentralized MARL algorithm for the state-of-the-art deep RL agent: advantage actor critic (A2C), within the context of ATSC. In particular, two methods are proposed to stabilize the learning procedure, by improving the observability and reducing the learning difficulty of each local agent. \bt{The proposed multi-agent A2C is compared against independent A2C and independent Q-learning algorithms, in both a large synthetic traffic grid and a large real-world traffic network of Monaco city, under simulated peak-hour traffic dynamics. Results demonstrate its optimality, robustness, and sample efficiency over other state-of-the-art decentralized MARL algorithms.}
\end{abstract}

\begin{IEEEkeywords}
Adaptive traffic signal control, Reinforcement learning, Multi-agent reinforcement learning, Deep reinforcement learning, Actor-critic.
\end{IEEEkeywords}


\section{Introduction}
As a consequence of population growth and urbanization, the transportation demand is steadily rising in the metropolises worldwide. The extensive routine traffic volumes bring pressures to existing urban traffic infrastructure, resulting in everyday traffic congestions. Adaptive traffic signal control (ATSC) aims for reducing potential congestions in saturated road networks, by adjusting the signal timing according to real-time traffic dynamics. \bt{Early-stage ATSC methods solve optimization problems to find efficient coordination and control policies. Successful products, such as SCOOT~\cite{hunt1982scoot} and SCATS~\cite{luk1984two}, have been installed in hundreds of cities across the world. OPAC~\cite{gartner1982demand} and PRODYN~\cite{henry1984prodyn} are similar methods, but their relatively complex computation makes the implementation less popular. Since the 90s, various interdisciplinary techniques have been applied to ATSC, such as fuzzy logic~\cite{gokulan2010distributed}, genetic algorithm~\cite{ceylan2004traffic}, and immune network algorithm~\cite{darmoul2017multi}.}

Reinforcement learning (RL), formulated under the framework of Markov decision process (MDP), is a promising alternative to learn ATSC based on real-world traffic measurements~\cite{sutton1998reinforcement}. Unlike traditional model-driven approaches, RL does not rely on heuristic assumptions and equations. Rather it directly fits a parametric model to learn the optimal control, based on its experience interacting with the complex traffic systems. Traditional RL fits simple models such as piece-wise constant table and linear regression (LR)~\cite{szepesvari2010algorithms}, leading to limited scalability or optimality in practice. \bt{Recently, deep neural networks (DNNs) have been successfully applied to enhance the learning capacity of RL on complex tasks~\cite{mnih2015human}.}

To utilize the power of deep RL, appropriate RL methods need to be adapted. There are three major methods: value-based, policy-based, and actor-critic methods. In value-based methods, such as {\it Q-learning}, the long-term state-action value function is parameterized and updated using step-wise \bt{experience}~\cite{watkins1992q}. Q-learning is off-policy\footnote{In off-policy learning, the behavior policy for sampling the experience is different from the target policy to be optimized.}, so it enjoys efficient updating with bootstrapped sampling of {\it experience replay}. \bt{However its update is based on one-step {\it temporal difference}, so the good convergence relies on a stationary MDP transition, which is less likely in ATSC.} As the contrast, in policy-based methods, such as \bt{REINFORCE}, the policy is directly parameterized and updated with sampled episode return~\cite{williams1992simple}. REINFORCE is on-policy so the transition can be nonstationary within each episode. \bt{The actor-critic methods further reduce the bias and variance of policy-based methods by using another model to parameterize the value function~\cite{konda1999actor}. A recent work has demonstrated that actor-critic outperforms Q-learning in ATSC with centralized LR agents~\cite{aslani2017adaptive}.} This paper focuses on the state-of-the-art  {\it advantage actor-critic} (A2C) where DNNs are used for both policy and value approximations~\cite{mnih2016asynchronous}.

Even though DNNs have improved the scalability of RL, training a centralized RL agent is still infeasible for large-scale ATSC. First, we need to collect all traffic measurements in the network and feed them to the agent as the global state. This centralized state processing itself will cause high latency and failure rate in practice, and the topological information of the traffic network will be lost. Further, the joint action space of the agent grows exponentially in the number of signalized intersections. 
Therefore, it is efficient and natural to formulate ATSC as a cooperative multi-agent RL (MARL) problem, where each intersection is controlled by a local RL agent, upon local observation and limited communication. MARL has a long history and has mostly focused on Q-learning, by distributing the global Q-function to local agents. One approach is to design a coordination rule based on the tradeoff between optimality and scalability~\cite{guestrin2002coordinated,kok2006collaborative}. A simpler and more common alternative is {\it independent} Q-learning (IQL)~\cite{tan1993multi}, in which each local agent learns its own policy independently, by modeling other agents as parts of the environment dynamics. IQL is completely scalable, but it has issue on convergence, since now the environment becomes more partially observable and nonstationary, as other agents update their policies. This issue was addressed recently for enabling experience replay in deep MARL~\cite{foerster2017stabilising}.

To the best of our knowledge, this is the first paper that formulates independent A2C (IA2C) for ATSC, by extending the idea of IQL on A2C. In order to develop a stable and robust IA2C system, two methods are further proposed to address the partially observable and nonstationary nature of IA2C, under limited communication. First, we include the observations and {\it fingerprints} of neighboring agents in the state, so that each local agent has more information regarding the regional traffic distribution and cooperative strategy. Second, we introduce a {\it spatial} discount factor to scale down the observation and reward signals of neighboring agents, so that each local agent focuses more on improving traffic nearby.  From the convergence aspect, the first approach increases the fitting power while the second approach reduces the fitting difficulty. We call the stabilized IA2C the multi-agent A2C (MA2C). \bt{MA2C is evaluated in both a synthetic large traffic grid and a real-world large traffic network, with delicately designed traffic dynamics for ensuring a certain difficulty level of MDP. Numerical experiments confirm that, MA2C outperforms IA2C and state-of-the-art IQL algorithms in robustness and optimality.} The code of this study is open sourced\footnote{See \url{https://github.com/cts198859/deeprl_signal_control}.}.


\section{Background} \label{sec:backgroud}
\subsection{Reinforcement Learning} \label{sec:rl_basic}
RL learns to maximize the long-term return of a MDP. In a fully observable MDP, the agent observes the true  state of the environment $s_t \in \scal$ at each time $t$, performs an action $u_t \in \ucal$ according to a policy $\pi(u|s)$. Then transition dynamics happens $s_{t+1} \sim p(\cdot|s_t, u_t)$, and an immediate step reward $r_t = r(s_t, u_t, s_{t+1})$ is received. In an infinite-horizon MDP, sampled total return under policy $\pi$ is $R^\pi_t = \sum_{\tau=t}^\infty \gamma^{\tau-t} r_\tau$, where $\gamma \in [0,1)$ is a discount factor. The expected total return is represented as its {\it Q-function} $Q^\pi(s,u) = \Expec [R^\pi_t|s_t=s, u_t=u]$. The optimal Q-function $Q^* = \max_\pi Q^\pi$ yields an optimal greedy policy $\pi^*(u|s): u \in \argmax_{u'} Q^*(s,u')$, and $Q^*$ is obtained by solving the Bellman equation $\tcal Q^* = Q^*$~\cite{bellman1957markovian}, with a {\it dynamic programming} (DP) operator $\tcal$:

\begin{equation}
\label{eq:tq}
\tcal Q(s,u) = r(s,u) + \gamma  \sum_{s'\in\scal} p(s'|s,u) \max_{u'\in\ucal} Q(s',u'),
\end{equation}

where $r(s,u) = \Expec_s' r(s,u,s')$ is the expected step reward. 
In practice, $r$ and $p$ are unknown to the agent so the above {\it planning} problem is not well defined. Instead, RL performs data-driven DP based on the sampled \bt{experience} $(s_t,u_t,s'_t,r_t)$.

 \subsubsection{Q-learning}
Q-learning is the fundamental RL method that fits the Q-function with a parametric model $Q_\theta$, such as Q-value table~\cite{watkins1992q}, LR~\cite{szepesvari2010algorithms}, or DNN~\cite{mnih2015human}. Given experience $(s_t,u_t,s'_t,r_t)$,  Eq.~\eqref{eq:tq} is used to estimate $\hat{\tcal} Q_{\theta^-}(s_t,u_t) = r_t + \gamma \max_{u'\in \ucal} Q_{\theta^-}(s'_t,u')$ using a frozen recent model $\theta^-$, then temporal difference $(\hat{\tcal} Q_{\theta^-} - Q_\theta) (s_t,u_t)$ is used to updated $\theta$.
A  \bt{behavior policy}, such as $\epsilon-$greedy, is used in Q-learning to explore rich experience to reduce the regression variance. Experience replay is applied in deep Q-learning to reduce the variance furthermore by sampling less correlated experience in each minibatch.

 \subsubsection{Policy Gradient}
Policy gradient (REINFORCE) directly fits the policy with a parameterized model $\pi_\theta$~\cite{sutton2000policy}. Each update of $\theta$ increases the likelihood for selecting the action that has the high ``optimality'', measured as the sampled total return. Thus the loss is
\begin{equation}
\label{eq:pg-loss}
\lcal(\theta) = -\frac{1}{|B|}\sum_{t \in B}  \log \pi_\theta(u_t|s_t) \hat{R}_t,
\end{equation}
where each minibatch $B=\{(s_t,u_t,s'_t,r_t)\}$ contains the experience trajectory, \ie,  $s'_t = s_{t+1}$, and each return is estimated as $\hat{R}_t= \sum_{\tau=t}^{t_B - 1} \gamma^{\tau-t} r_\tau$, here $t_B$ is the last step in minibatch. 
Policy gradient does not need a behavior policy since $\pi_\theta(u|s)$ naturally performs exploration and exploitation. Further, it is robust to nonstationary transitions within each trajectory since it directly uses return instead of estimating it recursively by $\hat{\tcal}$. However it suffers from high variance as $\hat{R}_t$ is much more noisy than fitted return $Q^{\pi_{\theta^-}}$.

\subsubsection{Advantage Actor-Critic}
A2C improves the policy gradient by introducing a {\it value} regressor $V_w $ to estimate $\Expec [R^\pi_t|s_t=s]$~\cite{konda1999actor}. First, it reduces the bias of sampled return by adding the value of the last state $R_{t} = \hat{R}_t + \gamma^{t_B-t}V_{w^-}(s_{t_B})$; Second, it reduces the variance of sampled return by using $A_t := R_{t} - V_{w^-}(s_t)$, which is interpreted as the sampled advantage $A^\pi(s,u) := Q^\pi(s,u) - V^\pi(s)$. Then Eq.~\eqref{eq:pg-loss} becomes
\begin{equation}\label{eq:p-loss}
\lcal(\theta) = -\frac{1}{|B|}\sum_{t \in B}  \log \pi_\theta(u_t|s_t) A_t.
\end{equation}
The loss function for value updating is:
\begin{equation}\label{eq:v-loss}
\lcal(w) = \frac{1}{2|B|}\sum_{t \in B}  \left(R_{t} - V_w(s_t)\right)^2.
\end{equation}

\subsection{Multi-agent Reinforcement Learning}
In ATSC, multiple signalized intersection agents participate a cooperative game to optimize the global network traffic objectives. Consider a multi-agent network $G(\vcal, \ecal)$, where each agent $i \in \vcal$ performs a discrete action $u_i \bt{\in \ucal_i}$, \bt{communicates to a neighbor via edge $ij \in \ecal$,} and shares the global reward $r(s, u)$. Then the joint action space is $\ucal = \times_{i \in \vcal} \ucal_i$, which makes centralized RL infeasible. MARL, mostly formulated in the context of Q-learning, distributes the global action to each local agent by assuming the global Q-function is decomposable $Q(s,u) = \sum_{i \in \vcal} Q_i(s, u)$. To simplify the notation, we assume each local agent can observe the global state, and we will relax this assumption for ATSC in Section~\ref{sec:ma2c} and \ref{sec:ma2c_traffic}.

Coordinated Q-learning is one MARL approach that performs iterative message passing or control syncing among neighboring agents to achieve desired tradeoff between optimality and scalability. In other words, $Q_i(s,u) \approx Q_i(s,u_i) + \sum_{j \in \vcal_i} M_j(s, u_j, u_{\vcal_j})$, where $\vcal_i$ is the neighborhood of agent $i$, and $M_j$ is the message from neighbor $j$, regarding the impact of $u_i \in u_{\vcal_j}$ on its local traffic condition. \cite{guestrin2001multiagent} applied variable elimination after passing Q-function as the message, while \cite{kok2006collaborative} proposed a max-plus message passing. This approach  (1) requires additional computation to obtain the coordinated control during execution, and (2) requires heuristics and assumptions to decompose Q-function and formulate message-passing, which can be potentially learned by IQL described shortly.

IQL is the most straightforward and popular approach, in which each local Q-function only depends on the local action, \ie, $Q_i(s, u) \approx Q_i(s, u_i)$~\cite{tan1993multi}. IQL is completely scalable, but without message passing, it suffers from partial observability and non-sationary MDP, because it implicitly formulates all other agents' behavior as part of the environment dynamics while their policies are continuously updated during training.
.
To address this issue, each local agent needs the information of other agents' policies. \cite{tesauro2004extending} included the policy network parameter of each other agent for fitting local Q-function, \ie, $Q_i(s,u) \approx Q_{\theta_i}(s,u_i,\theta_{-i})$, while \cite{foerster2017stabilising} included low-dimensional fingerprints, \ie, $Q_i(s,u) \approx Q_{\theta_i}(s, u_i, x_{-i})$. To handle the nonstationary transition in experience replay, {\it importance sampling} is applied to estimate the temporal difference of other agents' policies between the sampled time $t$ and the updating time $\tau$, as $\pi_{\tau, -i}(u_{-i}|s) / \pi_{t,-i}(u_{-i}|s)$, and to adjust the batch loss. MARL with A2C has not been formally addressed in literature, and will be covered in Section~\ref{sec:ma2c}.

\section{Related Work}  \label{sec:related_work}
The implementation of RL has been extensively studied in ATSC. Tabular Q-learning was the first RL algorithm applied, at an isolated intersection \cite{wiering2004intelligent}. Later, LR Q-learning was adapted for scalable fitting over continuous states. \cite{cai2009adaptive} and \cite{prashanth2011reinforcement} designed heuristic state features, while \cite{chu2015traffic} integrated macroscopic fundamental diagram to obtain more informative features. However, LR was too simple to capture the Q-function under complex traffic dynamics. Thus kernel method was applied to extract nonlinear features from low-dimensional states~\cite{chu2014kernel}. Kernel method was also applied in LR actor-critic recently, under realistically simulated traffic environments~\cite{aslani2017adaptive}. Alternatively, natural actor-critic was applied to improve the fitting accuracy of LR in ATSC~\cite{richter2007natural}. Deep RL was implemented recently, but most of them had impractical assumptions or oversimplified traffic environments. \cite{chu2016large} and \cite{casas2017deep} verified the superior fitting power of deep Q-learning and deep deterministic policy gradient, respectively, under simplified traffic environment. \cite{genders2016using} applied deep Q-learning in a more realistic traffic network, but with infeasible microscopic discrete traffic state. \bt{\cite{genders2018evaluating} explored A2C with different types of state information.} Due to the scalability issue, most centralized RL studies conducted experiments in either isolated intersections or small traffic networks.

Despite rich history of RL, only a few studies have addressed MARL in ATSC, and most of them have focused on Q-learning. \cite{wiering2000multi} applied model-based tabular IQL to each intersection while \cite{chu2016rrl} extended LR IQL to dynamically clustered regions to improve observability. \bt{\cite{aziz2018learning} studied LR IQL and its on-policy version independent SARSA learning, with observability improved by neighborhood information sharing.} \cite{van2016coordinated} applied deep IQL, with the partial observability addressed with {\it transfer planning}, but the states were infeasible and the simulated traffic environments were oversimplified in that study. Alternatively, coordinated Q-learning was implemented with various message-passing methods. \cite{el2013multiagent} designed a heuristic neighborhood communication for tabular Q-learning agents, where each message contained the estimated neighbor policies. \bt{\cite{zhu2015junction} proposed a junction-tree based hierarchical message-passing rule to coordinate tabular Q-learning agents.} \cite{chu2017traffic} applied the max-sum communication for LR Q-learning agents, where each message indicated the impact of a neighbor agent on each local Q-value.

To summarize, traditional RL has been widely applied in ATSC, and some works have proposed realistic state measurements and decent traffic dynamics based on insightful domain-specific knowledge. However, benchmark traffic environments are still missing for fair comparisons across proposed algorithms. On the other hand, there is still no comprehensive and realistic studies proposed for implementing deep RL in practical ATSC. MARL has been addressed in a few works, mostly under the context of Q-learning.

\section{Multi-agent Advantage Actor-Critic}\label{sec:ma2c}
MARL is mostly addressed in the context of Q-learning. In this section, we first formulate IA2C by extending the observations of IQL to the actor-critic method. Furthermore, we propose two approaches to stabilizing IA2C as MA2C. The first approach is inspired from the works for stabilizing IQL~\cite{tesauro2004extending,foerster2017stabilising}, where the recent policies of other agents are informed to each local agent. The second approach proposes a novel spatial discount factor to scale down the signals from agents far away. In other words, each local value function is smoother and more correlated to local states, and each agent focuses more on improving local traffic condition. As a result, the convergence becomes more stable even under limited communication and partial observation. This section proposes a general MA2C framework under limited neighborhood communication, while its implementation details for ATSC will be described in Section~\ref{sec:ma2c_traffic}.

\subsection{Independent A2C}
In a multi-agent network $G(\vcal, \ecal)$, $i$ and $j$ are neighbors if there is an edge between them. The neighborhood of $i$ is denoted as $\ncal_i$ and the local region is $\vcal_i =  \ncal_i \cup {i}$. Also, the distance between any two agents $d(i,j)$ is measured as the minimum number of edges connecting them. For example, $d(i,i) = 0$ and $d(i,j) = 1$ for any $j \in \ncal_i$. In IA2C, each agent learns its own policy $\pi_{\theta_i}$ and the corresponding value function $V_{w_i}$.

We start with a strong assumption where the global reward and state are shared among agents. Then centralized A2C updating can be easily extended to IA2C, by estimating local return as

\begin{equation}
\label{eq:R_ia2c}
R_{t,i} = \hat{R}_t + \gamma^{t_B-t} V_{w_i^-} (s_{t_B} | \pi_{\theta^-_{-i}}).
\end{equation}
The value gradient $\nabla \lcal(w_i)$ is consistent since $ \hat{R}_t$ is sampled from the same stationary policy $\pi_{\theta^-}$. To obtain policy gradient $\nabla \lcal(\theta_i)$,  $V_{w_i}: \scal \times \ucal_i \to \reals$ is served as the estimation of marginal impact of $\pi_{\theta_i}$ on the future return.  However, if each agent follows Eq.~\eqref{eq:R_ia2c}, each value gradient will be identical towards the global value function $V^{\pi}$ instead of the local one $V^{\pi_i} = \Expec_{\pi_{-i}} V^{\pi}$. As far as $\theta_{-i}$ is fixed, $\pi_{\theta_i}$ will still converge to the best according policy under this updating, and optimal policy $\pi_{\theta^*_i}$ can be achieved if $\theta_{-i} = \theta^*_{-i}$. However, when $\theta_{-i}$ is actively updated, the policy gradient may be inconsistent across minibatches, since the advantage is conditioned on changing $\pi_{\theta_{-i}}$, even it is stationary per trajectory.

Global information sharing is infeasible in real-time ATSC due to the communication latency, so we assume the communication is limited to each local region. In other words, local policy and value regressors take $s_{t, \vcal_i} := \{s_{t,j}\}_{j \in \vcal_i}$ instead of $s_t$ as the input state. Global reward is still allowed since it is only used in offline training. Eq.~\eqref{eq:R_ia2c} is valid here, by replacing the value estimation of the last state with $V_{w_i^-} (s_{t_B, \vcal_i} | \pi_{\theta^-_{-i}})$. Then the value loss Eq.~\eqref{eq:v-loss} becomes

\begin{equation}
\label{eq:v-loss_ia2c}
\lcal(w_i) = \frac{1}{2|B|}\sum_{t \in B}  \left(R_{t, i} - V_{w_i}(s_{t, \vcal_i})\right)^2.
\end{equation}
Clearly, $V_{w_i}$ suffers from partial observability, as $s_{t, \vcal_i}$ is a subset of $s_t$ while $\Expec R_{t, i}$ depends on $s_t$. Similarly, the policy loss Eq.~\eqref{eq:p-loss} becomes

\begin{equation}
\label{eq:p-loss_ia2c}
\lcal(\theta_i) = -\frac{1}{|B|}\sum_{t \in B}  \log \pi_{\theta_i}(u_{t,i}|s_{t, \vcal_i}) A_{t,i},
\end{equation}
where $A_{t,i} = R_{t, i} - V_{w_i^-}(s_{t, \vcal_i})$. The nonstaionary updating issue remains, since $R_{t, i}$ is conditioned on the current policy $\pi_{\theta^-_{-i}}$, while both $\theta^-_i$ and $w^-_i$ are updated under the previous policy $\pi_{\theta'_{-i}}$. This inconsistency effect can be mitigated if each local policy updating is smooth, \ie, the KL divergence $D_{KL}(\pi_{\theta^-_{-i}} || \pi_{\theta_{-i}})$ is small enough, but it will slow down the convergence. Partial observability also exists as $\pi_{\theta_i}(\cdot|s_{t, \vcal_i})$ cannot fully capture the impact of $\hat{R}_{t}$.

\subsection{Multi-agent A2C}
In order to stabilize IA2C convergence and enhance its fitting power, we propose two approaches. First, we include information of neighborhood policies to improve the observability of each local agent. Second, we introduce a spatial discount factor to weaken the state and reward signals from other agents. In IQL, additional information is included to represent the other agents' behavior policies.\cite{tesauro2004extending} directly includes the Q-value network parameters $\theta^-_{-i}$, while \cite{foerster2017stabilising} includes low-dimensional fingerprints, such as $\epsilon$ of the $\epsilon$-greedy exploration and the number of updates so far. Fortunately, the behavior policy is explicitly parameterized in A2C, so a natural approach is including probability simplex of policy $\pi_{\theta^-_{-i}}$. Under limited communication, we include sampled latest policies of neighbors $\pi_{t-1,\ncal_i} = [\pi_{t-1,j}]_{j \in \ncal_i}$ in the DNN inputs, besides the current state $s_{t, \vcal_i}$. The sampled local policy is calculated as

\begin{equation}
\label{eq:pi_ma2c}
\pi_{t,i} = \pi_{\theta^-_i}(\cdot|s_{t, \vcal_i}, \pi_{t-1,\ncal_i}).
\end{equation}
Rather than long-term neighborhood behavior, the real-time recent neighborhood policy is informed to each local agent. This is based on two ATSC facts: 1) the traffic state is changing slowly in short windows, so the current step policy is very similar to last step policy. 2) the traffic state dynamics is Markovian, given the current state and policy.

Even if the local agent knows the local region state and the neighborhood policy, it is still difficult to fit the global return by local value regressor. To relax the global cooperation, we assume the global reward is decomposable as $r_t = \sum_{i \in \vcal} r_{t,i}$, which is mostly valid in ATSC. Then we introduce a spatial discount factor $\alpha$ to adjust the global reward for agent $i$ as

\begin{equation}
\label{eq:r_ma2c}
\tilde{r}_{t,i} = \sum_{d=0}^{D_i} \left( \sum_{j \in \vcal | d(i, j) =d}  \alpha^d r_{t,j}\right),
\end{equation}
where $D_i$ is the maximum distance from agent $i$. Note $\alpha$ is similar to the temporal discount factor $\gamma$ in RL, rather it scales down the signals in spatial order instead of temporal order. Compared to sharing the same global reward across agents, the discounted global reward is more flexible for the trade-off between greedy control ($\alpha = 0$) and cooperative control ($\alpha=1$), and is more relevant for estimating the ``advantage'' of local policy $\pi_{\theta_i}$. Similarly, we use $\alpha$ to discount the neighborhood states as

\begin{equation}
\tilde{s}_{t,\vcal_i} = [s_{t,i}] \cup \alpha [s_{t,j}]_{j \in \ncal_i}.
\end{equation}

Given the discounted global reward, we have $\hat{R}_{t,i} = \sum_{\tau=t}^{t_B - 1} \gamma^{\tau-t} \tilde{r}_{\tau, i}$, and the local return Eq.~\eqref{eq:R_ia2c} becomes

\begin{equation}
\label{eq:R_ma2c}
\tilde{R}_{t,i} = \hat{R}_{t,i} + \gamma^{t_B-t} V_{w_i^-} (\tilde{s}_{t_B, \vcal_i}, \pi_{t_B-1,\ncal_i} | \pi_{\theta^-_{-i}}).
\end{equation}
The value loss Eq.~\eqref{eq:v-loss_ia2c} becomes

\begin{equation}
\label{eq:v-loss_ma2c}
\lcal(w_i) = \frac{1}{2|B|}\sum_{t \in B}  \left(\tilde{R}_{t,i} - V_{w_i}(\tilde{s}_{t, \vcal_i}, \pi_{t-1,\ncal_i})\right)^2.
\end{equation}
This updating is more stable since (1) additional fingerprints $\pi_{t-1,\ncal_i}$ are input to $V_{w_i}$ for  fitting $\pi_{\theta^-_{-i}}$ impact, and (2) spatially discounted return $\tilde{R}_{t,i}$ is more correlated to local region observations $(\tilde{s}_{t, \vcal_i}, \pi_{t-1,\ncal_i})$. The policy loss Eq.~\eqref{eq:p-loss_ma2c} becomes

\begin{align}
\label{eq:p-loss_ma2c}
\lcal(\theta_i) =& -\frac{1}{|B|}\sum_{t \in B} \bigg( \log \pi_{\theta_i}(u_{t,i}|\tilde{s}_{t, \vcal_i}, \pi_{t-1,\ncal_i}) \tilde{A}_{t,i}  \nonumber\\
&  - \beta \sum_{u_i \in \ucal_i} \pi_{\theta_i}\log \pi_{\theta_i} (u_{i}|\tilde{s}_{t, \vcal_i}, \pi_{t-1,\ncal_i}) \bigg)
\end{align}
where $\tilde{A}_{t,i} = \tilde{R}_{t, i} - V_{w_i^-}(\tilde{s}_{t, \vcal_i}, \pi_{t-1,\ncal_i})$, and the additional regularization term is the entropy loss of policy $\pi_{\theta_i}$ for encouraging the early-stage exploration. This new advantage emphasizes more on the marginal impact of $\pi_{\theta_i}$ on traffic state within local region.

Algorithm~\ref{algo:ma2c} illustrates the MA2C algorithm under the synchronous updating. $T$ is the planning horizon per episode, $|B|$ is the minibatch size, $\eta_w$ and $\eta_\theta$ are the learning rates for critic and actor networks. First, each local agent collects the experience by following the current policy, until enough samples are collected for minibatch updating (lines 3 to 8). If the episode is terminated in the middle of minibatch collection, we simply restart a new episode (lines 9 to 11). However the termination affects the return estimation. If $T$ is reached in the middle, the trajectory return (line 14) should be adjusted as

\begin{equation}
\label{eq:R_algo}
\hat{R}_{t,i}  = \begin{cases}
\sum_{\tau=t}^{T-1} \gamma^{\tau-t} \tilde{r}_{\tau, i} &\text{before reset,} \\
\sum_{\tau=t}^{t_B-1} \gamma^{\tau-t} \tilde{r}_{\tau, i} &\text{after reset.}
\end{cases}
\end{equation}
If $T$ is reached at the end, $\tilde{R}_{t,i} = \hat{R}_{t,i}$ (line 15), without the value estimation on future return in Eq.~\eqref{eq:R_ma2c}. Next, the minibatch gradients are applied to update each actor and critic networks (lines 16, 17), with constant or adaptive learning rates. Usually the first order gradient optimizers are used, such as stochastic gradient descent, RMSprop, and Adam. Finally, the training process is terminated if the maximum step is reached or a certain stop condition is triggered.

\begin{figure}[!h]
\removelatexerror
\begin{algorithm}[H]
\caption{Synchronous multi-agent A2C}
\label{algo:ma2c}
\SetAlgoLined
\Parameter{$\alpha$, $\beta$, $\gamma$, $T$, $|B|$, $\eta_w$, $\eta_\theta$.}
\KwResult{$\{w_i\}_{i \in \vcal}, \{\theta_i\}_{i \in \vcal}$.}
{\bf initialize} $s_0$, $\pi_{-1}$, $t \gets 0$, $k \gets 0$, $B = \emptyset$\;
\Repeat{Stop condition reached}{
	\tcc{explore experience}
	\For{$i \in \vcal$}{
		{\bf sample} $u_{t, i}$ from $\pi_{t, i}$ (Eq.~\eqref{eq:pi_ma2c})\;
		{\bf receive} $\tilde{r}_{t,i}$ (Eq.~\eqref{eq:r_ma2c}) and $\tilde{s}_{t,i}$\;
	}
	$B \gets B \cup \{(t, \tilde{s}_{t,i}, \pi_{t,i}, u_{t,i}, \tilde{r}_{t,i}, \tilde{s}_{t+1,i})\}_{i\in\vcal}$\;
	$t \gets t+1$, $k \gets k + 1$\;
	\tcc{restart episode}
	\If{$t = T$}{
		{\bf initialize} $s_0$, $\pi_{-1}$, $t \gets 0$\;
	}
	\tcc{update A2C}
	\If{$k = |B|$}{
		\For{$i \in \vcal$}{
			{\bf estimate} $\hat{R}_{\tau,i}$,  $\forall \tau \in B$\;
			{\bf estimate} $\tilde{R}_{\tau,i}$,  $\forall \tau \in B$\;
			{\bf update} $w_i$ with $\eta_w \nabla \lcal(w_i)$ (Eq.~\eqref{eq:v-loss_ma2c})\;
			{\bf update} $\theta_i$ with $\eta_\theta \nabla \lcal(\theta_i)$ (Eq.~\eqref{eq:p-loss_ma2c})\;
		}
		$B \gets \emptyset$, $k \gets 0$\;
	}
}
\end{algorithm}
\end{figure}


\section{MA2C for Traffic Signal Control} \label{sec:ma2c_traffic}
This section describes the implementation details of MA2C for ATSC, under the microscopic traffic simulator SUMO\cite{SUMO2012}. Specifically, we address the action, state, reward definitions, the A2C network structures and normalizations, the A2C training tips, and the evaluation metrics.

\subsection{MDP Settings}
\bt{Consider a simulated traffic environment over a period of $T_s$ seconds, we define $\Delta t$ as the interaction period between RL agents and the traffic environment, so that the environment is simulated for $\Delta t$ seconds after each MDP step. If $\Delta t$ is too long, RL agents will not be adaptive enough. If $\Delta t$ is too short, the RL decision will not be delivered on time due to computational cost and communication latency. Further, there will be safety concerns if RL control is switched too frequently. To further guarantee the safety, a yellow time $t_y < \Delta t$ is enforced after each traffic light switch. A recent work suggested $\Delta t = 10$s, and $t_y = 5$s~\cite{aslani2017adaptive}. In this paper, we set $\Delta t = 5$s, $t_y=2$s to ensure more adaptiveness, resulting in a planning horizon of $T=T_s/\Delta t$ steps.}

\subsubsection{Action definition}
\bt{There are several standard action definitions, such as phase switch~\cite{el2013multiagent},  phase duration~\cite{aslani2017adaptive}, and phase itself~\cite{prashanth2011reinforcement}. We follow the last definition and simply define each local action as a possible phase, or red-green combinations of traffic lights at that intersection. This enables more flexible and direct ATSC by RL agents. Specifically, we pre-define $\ucal_i$ as a set of all feasible phases for each intersection, and RL agent selects one of them to be implemented for a duration of $\Delta t$ at each step.}

\begin{myverbbox}{\wave}wave\end{myverbbox}
\begin{myverbbox}{\wait}wait\end{myverbbox}
\begin{myverbbox}{\queue}queue\end{myverbbox}
\begin{myverbbox}{\ild}laneAreaDetector\end{myverbbox}

\subsubsection{State definition}
\bt{
After combining the ideas of \cite{prashanth2011reinforcement} and \cite{aslani2017adaptive}, we define the local state as
\begin{align}
s_{t,i} &= \{\wait_t[l], \wave_t[l] \}_{ji \in \ecal, l \in L_{ji}},
\end{align}
where $l$ is each incoming lane of intersection $i$. $\wait$ [s] measures the cumulative delay of the first vehicle, while $\wave$ [veh] measures the total number of approaching vehicles along each incoming lane, within 50m to the intersection. Both $\wait$ and $\wave$ can be obtained by near-intersection induction-loop detectors (ILD), which ensures real-time ATSC. To simplify the implementation, we use $\ild$ in SUMO to collect the state information.}

\subsubsection{Reward definition}
Various objectives are selected for ATSC, but an appropriate reward of MARL should be spatially decomposable and frequently measurable. \bt{Here we refine the definition in \cite{prashanth2011reinforcement} as
\begin{equation}
\label{eq:r_atsc}
r_{t,i} = - \sum_{ji \in \ecal, l \in L_{ji}}(\queue_{t+\Delta t }[l] + a \cdot \wait_{t+\Delta t }[l]),
\end{equation}
where $a$ [veh/s] is a tradeoff coefficient, and $\queue$ [veh] is the measured queue length along each incoming lane. Notably the reward is post-decision so both $\queue$ and $\wait$ are measured at time $t+\Delta t$. We prefer this definition over others like cumulative delay~\cite{el2013multiagent} and wave~\cite{aslani2017adaptive}, since this reward is directly correlated to state and action, and emphasizes both traffic congestion and trip delay.}

\subsection{DNN Settings}
\subsubsection{Network architecture}
\begin{figure}[!h]
  \centering
  \includegraphics[width=.35\textwidth]{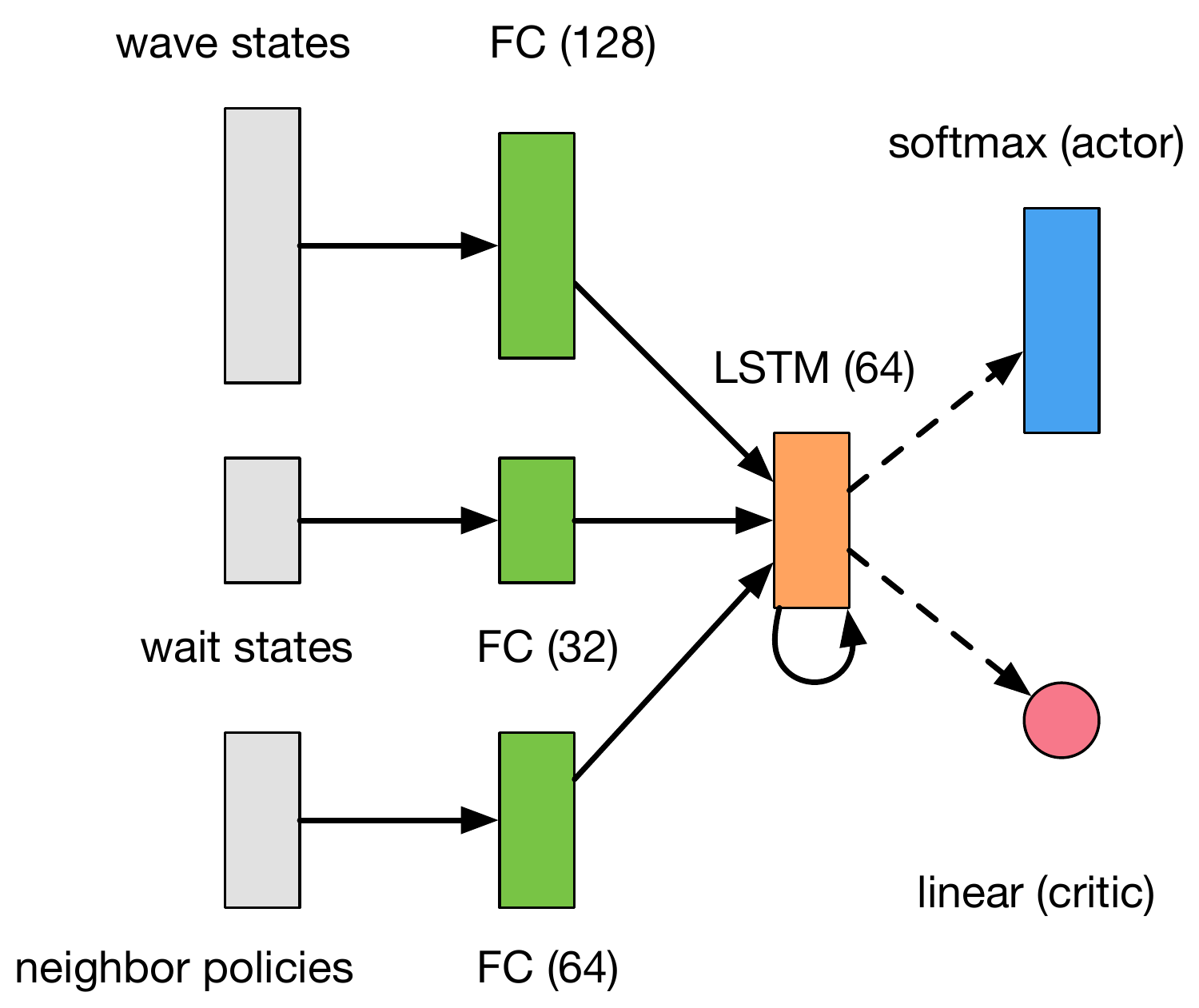}
  \caption{\bt{Proposed DNN structures of MA2C in ATSC. Hidden layer size is indicated inside parenthesis.}}
  \label{fig:dnn}
\end{figure}

In practice, the traffic flows are complex spatial-temporal data, so the MDP may become nonstationary if the agent only knows the current state. A straightforward approach is to input all historical states to A2C, but it increases the state dimension significantly and may reduce the attention of A2C on recent traffic condition. Fortunately, {\it long-short term memory} (LSTM) is a promising DNN layer that maintains hidden states to memorize short history~\cite{hochreiter1997long}. Thus we apply LSTM as the \bt{last hidden} layer to extract representations from different state types. Also, we train actor and critical DNNs separately, instead of sharing lower layers between them. Fig.~\ref{fig:dnn} illustrates the DNN structure, \bt{where $\wave$, $\wait$, and neighbor policies are first processed by separate fully connected (FC) layers. Then all hidden units are combined and input to a LSTM layer.} The output layer is softmax for actor and linear for critic. For DNN training, we use the state-of-the-art orthogonal initializer~\cite{saxe2013exact} and RMSprop as the gradient optimizer.

\subsubsection{Normalization}
Normalization is important for training DNN. \bt{For each of $\wave$ and $\wait$ states,  we run a greedy policy to collect the statistics for a certain traffic environment}, and use it to obtain an appropriate normalization. To prevent gradient explosion, all normalized states are clipped to [0, 2], \bt{and each gradient is capped at 40. Similarly, we normalize the reward and clip it to [-2, 2] to stabilize the minibatch updating.}

\section{Numerical Experiments} \label{sec:exp}
MARL based ATSC is evaluated in two SUMO-simulated traffic environments: \bt{a $5 \times 5$ synthetic traffic grid, and a real-world 30-intersection traffic network extracted from Monaco city~\cite{MoSTCodeca2018}
, under time-variant traffic flows. This section aims to design challenging and realistic traffic environments for interesting and fair comparisons across controllers.}

\subsection{General Setups}
\bt{
To demonstrate the efficiency and robustness of MA2C, we compare it to several state-of-the-art benchmark controllers. IA2C is the same as MA2C except the proposed stabilizing methods, so it follows the updating rules Eq.~\eqref{eq:v-loss_ia2c} and Eq.~\eqref{eq:p-loss_ia2c}. IQL-LR is LR based IQL, which can be regarded as the decentralized version of \cite{prashanth2011reinforcement}. IQL-DNN is the same as IQL-LR but uses DNN for fitting the Q-function. To ensure fair comparison, IQL-DNN has the same network structure, except the LSTM layer is replaced by a FC layer. This is because Q-learning is off-policy and it becomes meaningless to use LSTM on randomly sampled history.  Finally, Greedy is a decentralized greedy policy that selects the phase associated with the maximum total green $\wave$ over all incoming lanes. All controllers have the same action space, state space, and interaction frequency.}

\bt{
We train all MARL algorithms over 1M steps, which is around 1400 episodes given episode horizon $T = 720$ steps. We then evaluate all controllers over 10 episodes. Different random seeds are used for generating different training and evaluation episodes, but the same seed is shared for the same episode. For MDP, we set $\gamma = 0.99$ and $\alpha = 0.75$; For IA2C and MA2C, we set $\eta_\theta = 5e-4$, $\eta_w = 2.5e-4$, $|B| = 120$, and $\beta = 0.01$ in Algorithm~\ref{algo:ma2c}; For IQL, we set the learning rate $\eta_\theta = 1e-4$, the minibatch size $|B| = 20$, and the replay buffer size $1000$. Note the replay buffer size has too be small due to the partial observability of IQL. Also, $\epsilon-$greedy is used as the behavior policy, with $\epsilon$ linearly decaying from 1.0 to 0.01 during the first half of training.}

\subsection{Synthetic Traffic Grid}
\subsubsection{Experiment settings}
\begin{figure}[!h]
  \centering
  \includegraphics[width=.4\textwidth]{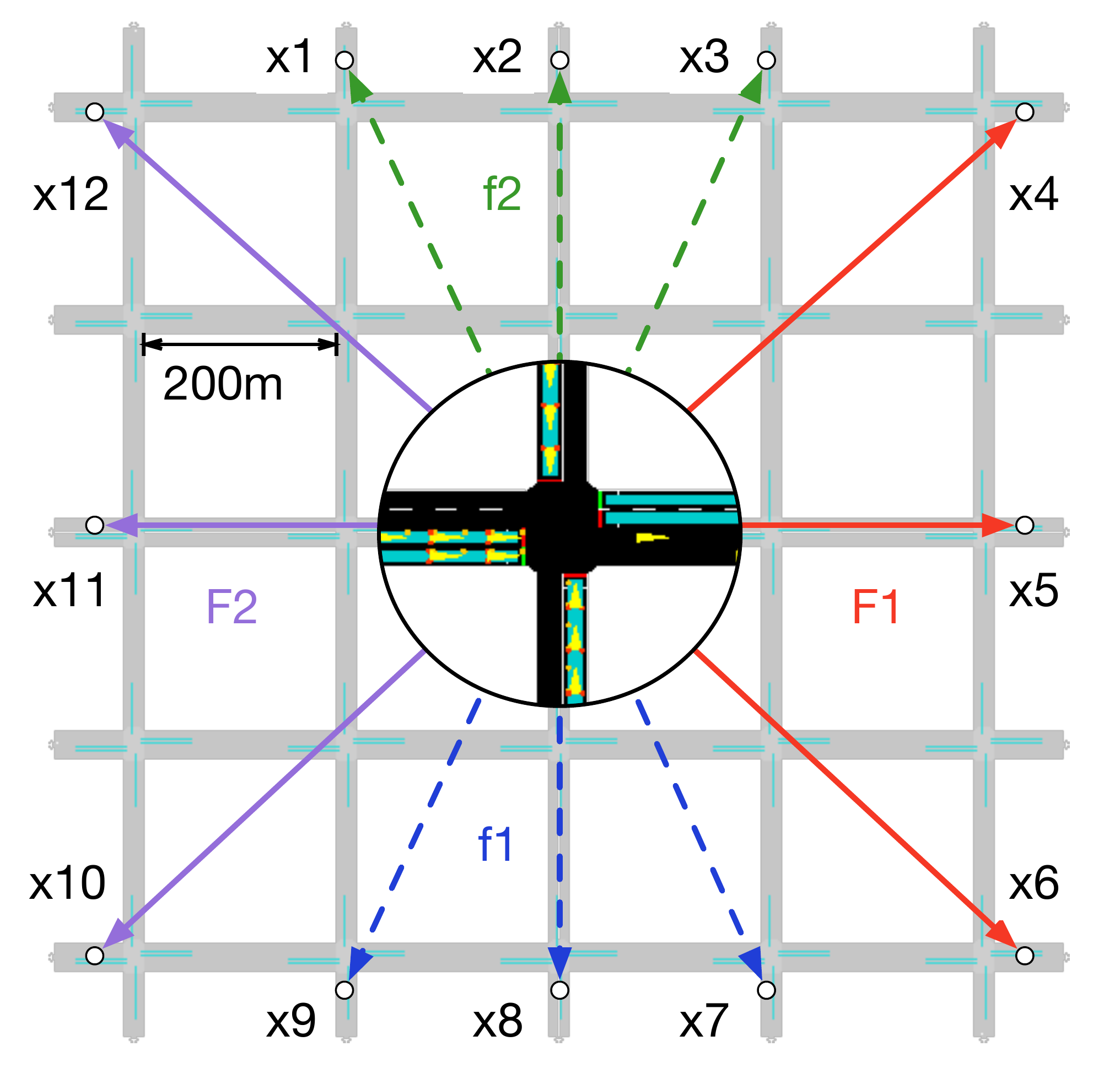}
  \caption{\bt{A traffic grid of 25 intersections, with an example intersection shown inside circle. Time variant major and minor traffic flow groups are shown as solid and dotted arrows.}}
  \label{fig:largenet}
\end{figure}

\begin{figure}[!h]
  \centering
  \includegraphics[width=.45\textwidth]{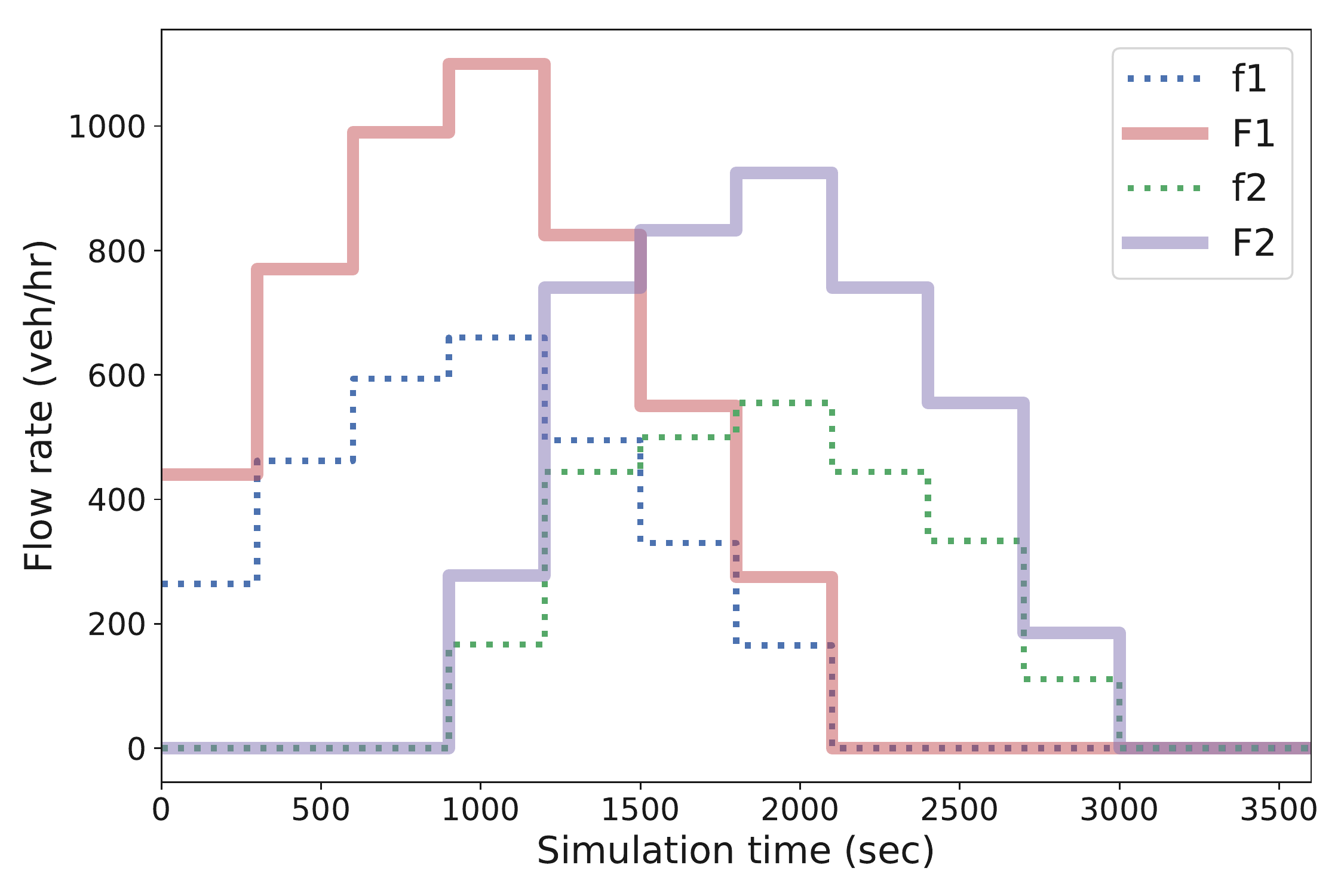}
  \caption{\bt{Traffic flows vs simulation time within the traffic grid.}}
  \label{fig:largeflow}
\end{figure}

\bt{The $5 \times 5$ traffic grid, as illustrated in Fig.~\ref{fig:largenet}, is formed by two-lane arterial streets with speed limit 20m/s and one-lane avenues with speed limit 11m/s. The action space of each intersection contains five possible phases: E-W straight phase, E-W left-turn phase, and three straight and left-turn phases for E, W, and N-S. Clearly, centralized RL agent is infeasible since the joint action space has the size of $5^{25}$. To make the MDP problem challenging, four time-variant traffic flow groups are simulated. At beginning, three major flows $F_1$ are generated with origin-destination (O-D) pairs $x_{10}$-$x_4$, $x_{11}$-$x_5$, and $x_{12}$-$x_6$, meanwhile three minor flows $f_1$ are generated with O-D pairs $x_1$-$x_7$, $x_2$-$x_8$, and $x_3$-$x_9$. After 15 minutes, the volumes of $F_1$ and $f_1$ start to decrease, while their opposite flows (with swapped O-D pairs) $F_2$ and $f_2$ start to be generated, as shown in Fig.~\ref{fig:largeflow}. Here the flows define the high-level traffic demand only, and the route of each vehicle is randomly generated during run-time. Regarding MDP settings, the reward coefficient $a$ is 0.2veh/s, and the normalization factors of $\wave$, $\wait$, and reward are  5veh, 100s, and 2000veh, respectively.}

\begin{figure}[!h]
  \centering
  \includegraphics[width=.45\textwidth]{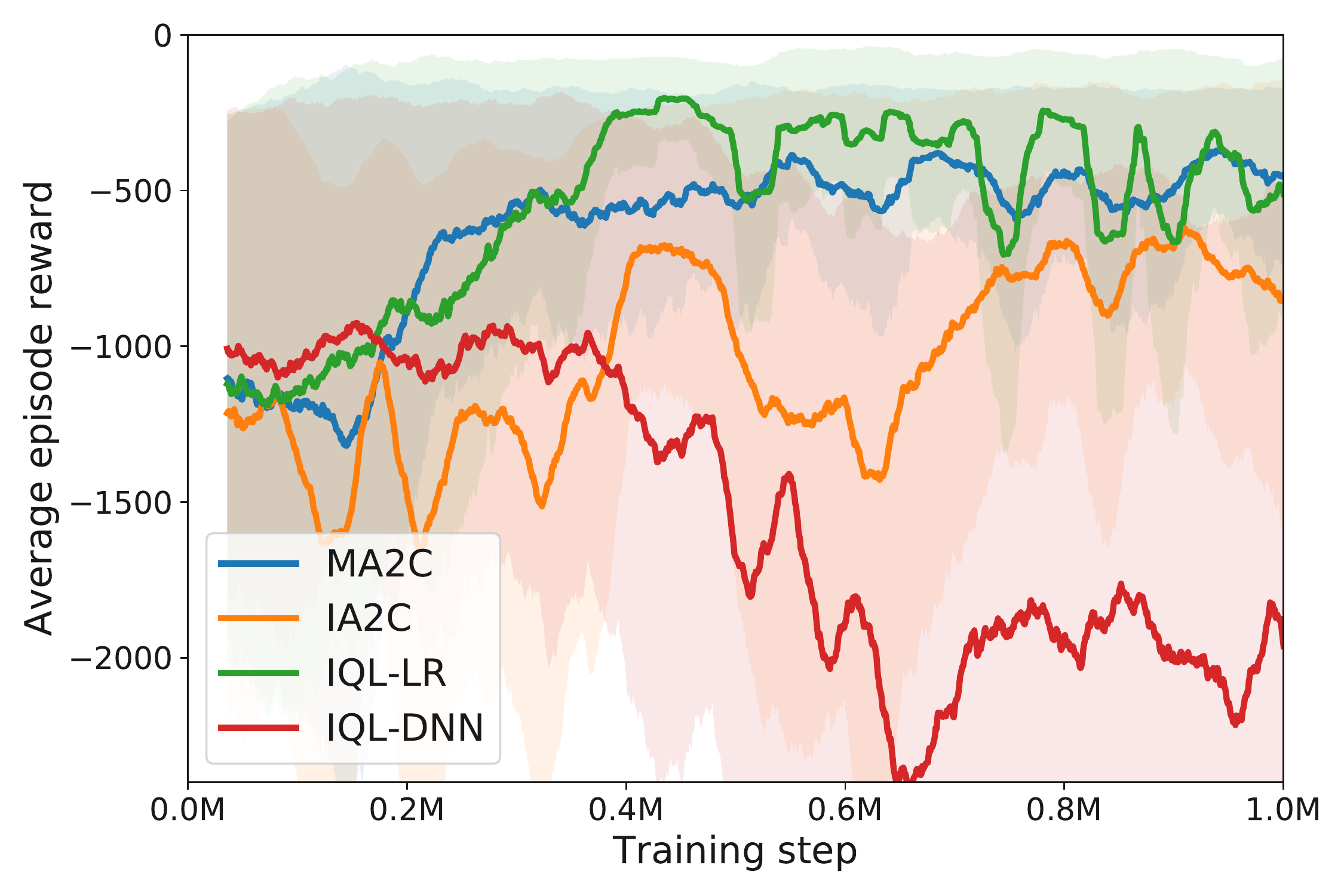}
  \caption{\bt{MARL training curves for synthetic traffic grid.}}
  \label{fig:large_train}
\end{figure}

\subsubsection{Training results}
\bt{Fig.~\ref{fig:large_train} plots the training curve of each MARL algorithm, where the solid line shows the average reward per training episode
\begin{equation}
\label{eq:eva_R}
\bar{R} = \frac{1}{T} \sum_{t=0}^{T-1} \left( \sum_{i \in \vcal} r_{t,i} \right),
\end{equation}
and the shade shows its standard deviation. 
Typically, training curve increases and then converges, as RL learns from cumulated experience and finally achieves a local optimum. In Fig.~\ref{fig:large_train},  IQL-DNN is failed to learn, while IQL-LR achieves a performance as good as MA2C does. This may because DNN overfits the Q-function using the partially observed states, misleading the exploitation when $\epsilon$ decreases. On the other hand, MA2C shows the best and the most robust learning ability, as its training curve steadily increases and then becomes stable with narrow shade.}

\begin{figure}[!h]
  \centering
  \includegraphics[width=.45\textwidth]{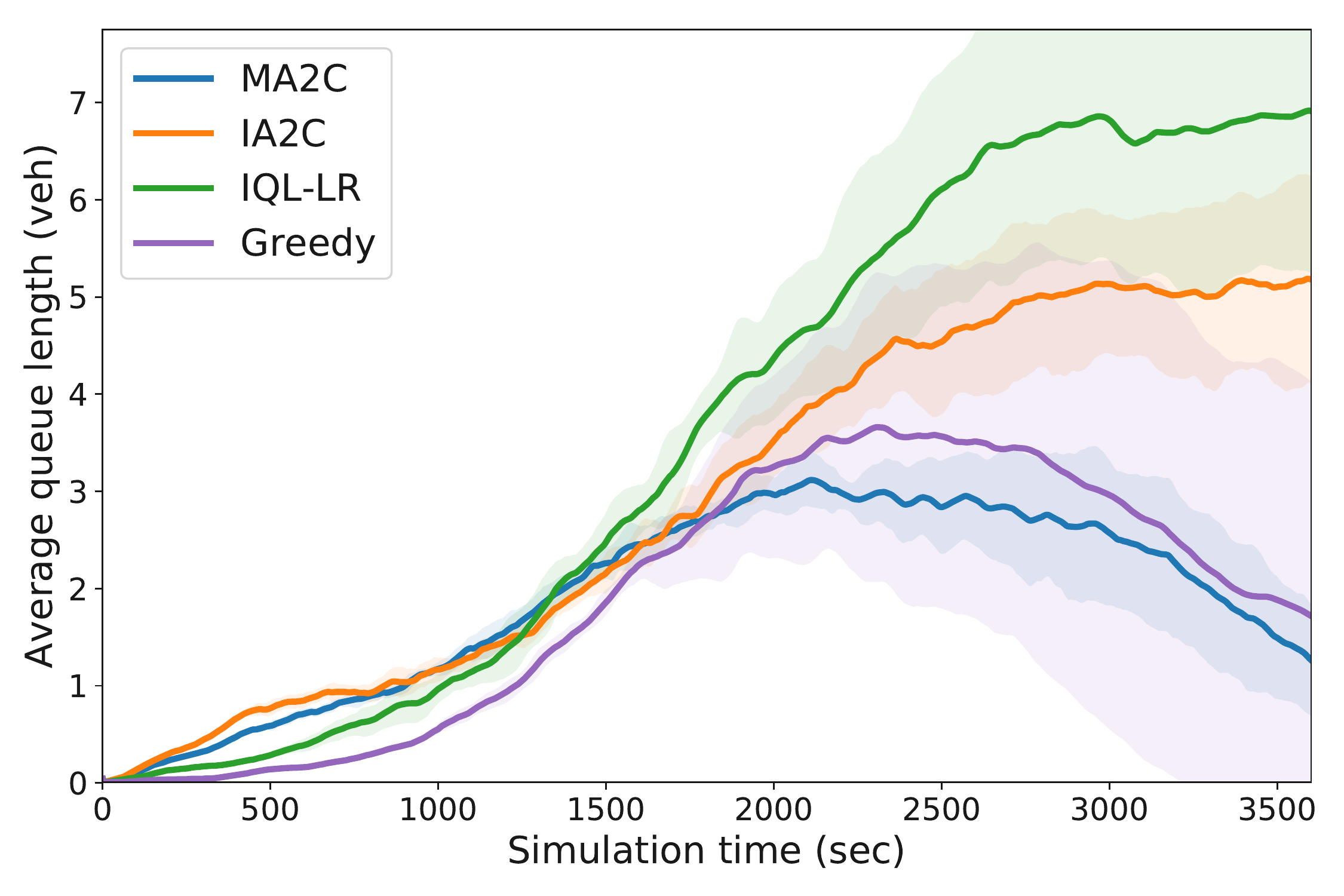}
  \caption{\bt{Average queue length in synthetic traffic grid.}}
  \label{fig:large_queue}
\end{figure}

\begin{figure}[!h]
  \centering
  \includegraphics[width=.45\textwidth]{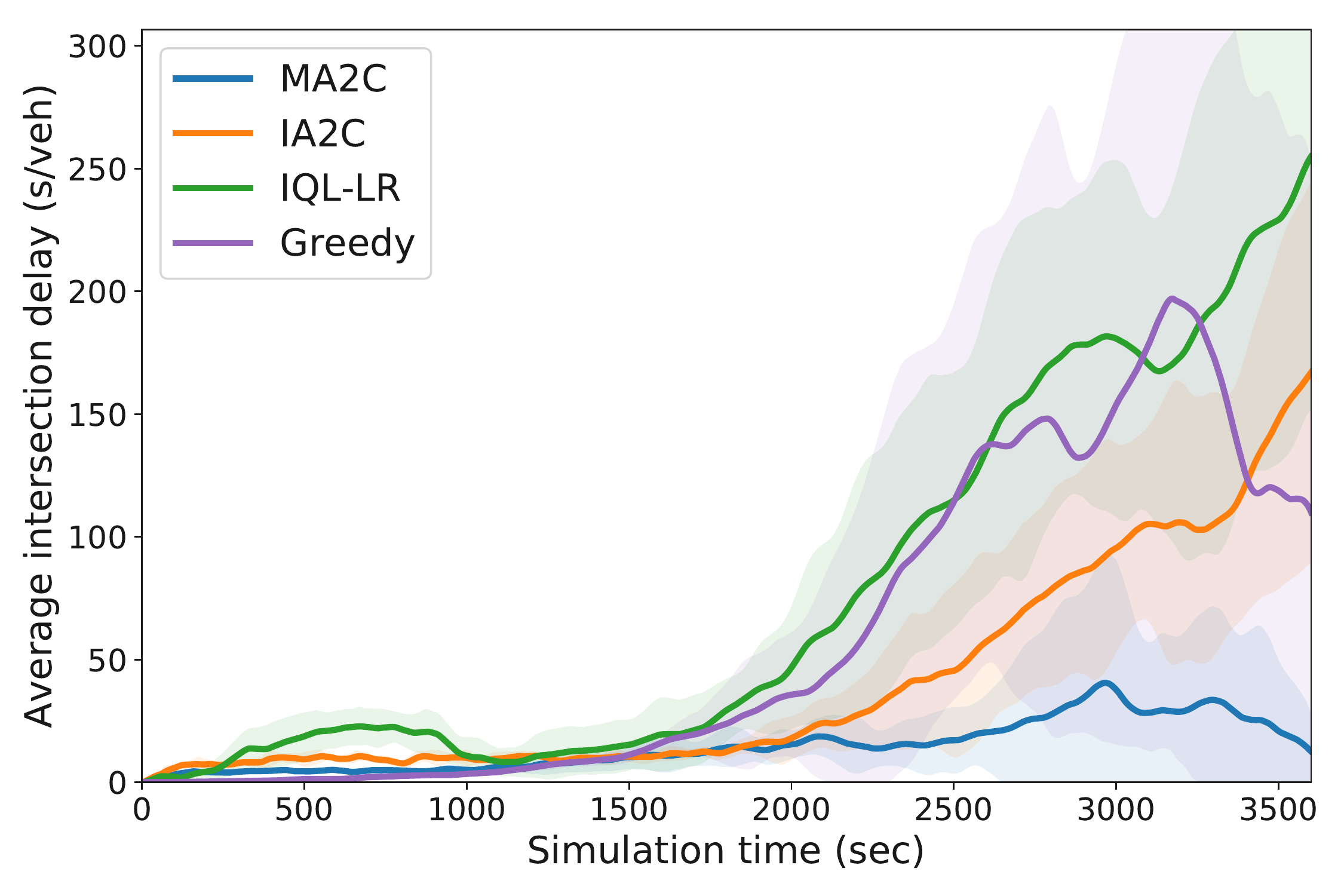}
  \caption{\bt{Average intersection delay in synthetic traffic grid.}}
  \label{fig:large_wait}
\end{figure}

\subsubsection{Evaluation results}
\bt{IQL-DNN is not included in evaluation as its policy is meaningless. The average $\bar{R}$ over evaluation are -414, -845, -972,    and -1409, for MA2C, IA2C, Greedy, and IQL-LR. Clearly MA2C outperforms other controllers for the given objective. Also, IQL-LR  is failed to beat IA2C in this evaluation over more episodes, which may due to the high variance in the learned policy.  Fig.~\ref{fig:large_queue} plots the average queue length of network at each simulation step, where the line shows the average and the shade shows the standard deviation across evaluation episodes. Both IQL-LR and IA2C fail to lean a sustainable policy to recover the congested network near the end. As contrast, a simple greedy policy does well for queue length optimization by maximizing the flow at each step, but its performance variation is high (see wide shade). MA2C leans a more stable and sustainable policy that achieves lower congestion level and faster recovery, by paying a higher queue cost when the network is less loaded at early stage.}

\bt{Fig.~\ref{fig:large_wait} plots the average intersection delay of network over simulation time. As expected, both IQL-LR and IA2C have monotonically increasing delays as they are failed to recover from the congestion. However, IA2C is able to maintain a more slowly increasing delay when the traffic grid is less saturated, so its overall performance is still better than the greedy policy, which does not explicitly optimize the waiting time. MA2C is able to maintain the delay at low level even at the peak.}

\begin{table*}[t]
\caption{ATSC performance in Monaco traffic network. Best values are in bold.}
\label{tab:realnet_perf}
\centering
\begin{tabular}{l | c c c c c | c c c c c}
\hline
\multirow{2}{*}{Metrics} & \multicolumn{5}{c}{Temporal Averages} & \multicolumn{5}{c}{Temporal Peaks}\\
 \cline{2-11}
& Greedy & MA2C & IA2C & IQL-LR & IQL-DNN & Greedy & MA2C & IA2C & IQL-LR & IQL-DNN\\
\hline
reward & -41.8 & \bf{-31.4} & -54.6 & -109.8  & -151.8 &  -86.4 & \bf{-78.7} &  -117.9 & -202.1 & -256.2\\
avg. queue length [veh] & 0.51 & \bf{0.29} & 0.52  & 1.19 & 1.57 & 1.08 & \bf{0.75} & 1.16 & 2.21 & 2.69\\
avg. intersection delay [s/veh] & 65.5 & \bf{23.5} & 60.7 & 100.3 &  127.0 &  272 & \bf{104} & 316 & 202 & 238\\
avg. vehicle speed [m/s] & 6.06 & \bf{6.81} & 5.36 & 4.04 & 2.07 &  \bf{14.96} & 14.26 & 14.26 & 14.26 & 13.98\\
trip completion flow [veh/s] & 0.54 & \bf{0.67} & 0.62 & 0.44  & 0.28 & 2.10 & \bf{2.40} &  2.10 & 1.60 & 1.20\\
trip delay [s] & 144 & \bf{114} & 165 & 207 & 360 & 2077 & \bf{1701} & 2418 & 2755 & 3283\\
\hline
\end{tabular}
\end{table*}

\subsection{Monaco Traffic Network}
\subsubsection{Experiment settings}
\begin{figure}[!h]
  \centering
  \includegraphics[width=.45\textwidth]{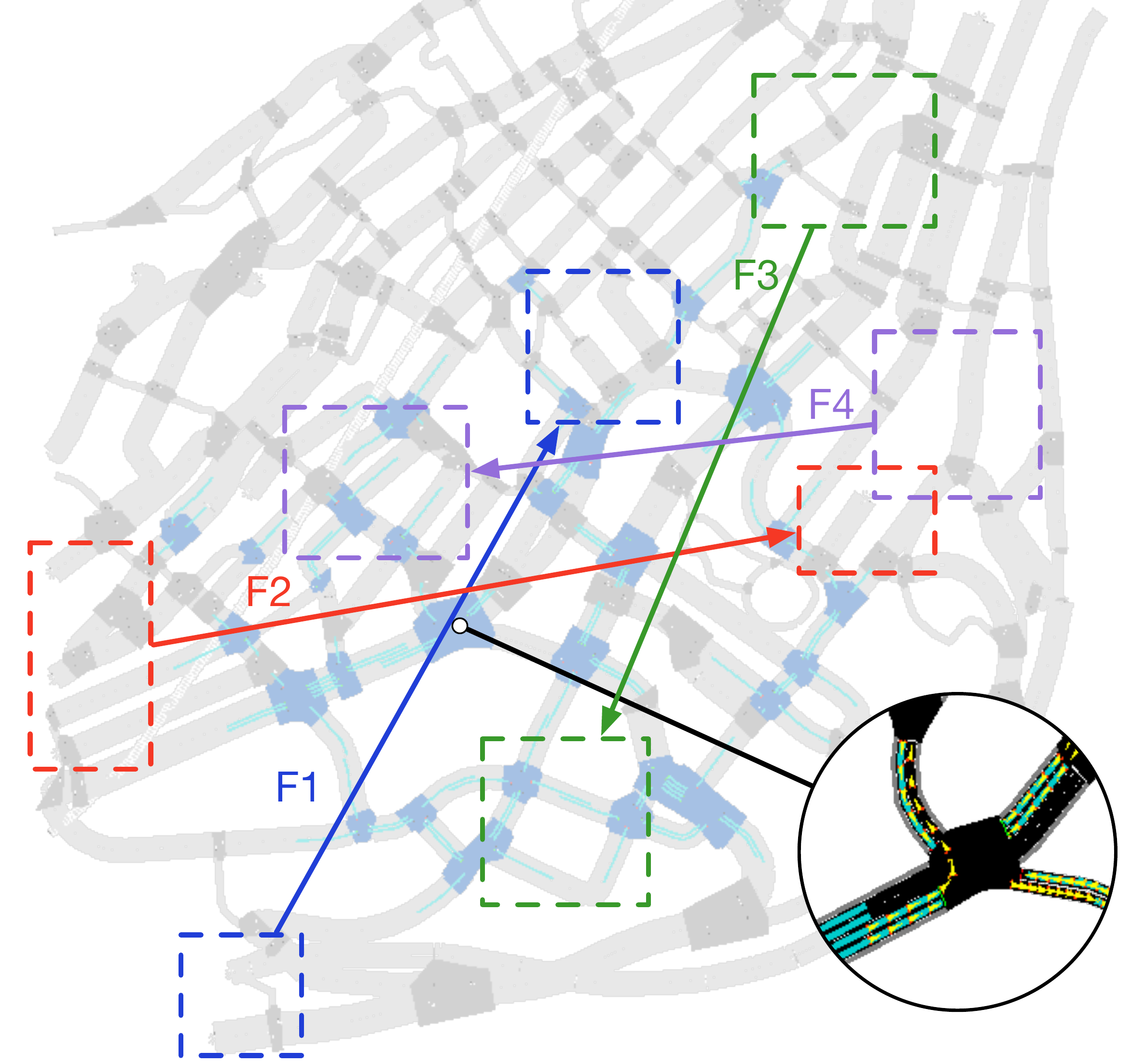}
  \caption{\bt{Monaco traffic network, with signalized intersections colored in blue. Four traffic flow groups are illustrated by arrows, with origin and destination inside rectangular areas.}}
  \label{fig:realnet}
\end{figure}

Fig.~\ref{fig:realnet} illustrates the studied area of Monaco city for this experiment, with a variety of road and intersection types. There are 30 signalized intersections in total: 11 are two-phase,  4 are three-phase, 10 are four-phase, 1 is five-phase, and the reset 4 are six-phase. \bbt{Further, in order to test the robustness and optimality of algorithms in challenging ATSC scenarios, intensive, stochastic, and time-variant traffic flows are designed to simulate the peak-hour traffic.} Specifically, four traffic flow groups are generated as a multiple of ``unit'' flows of 325veh/hr, with randomly sampled O-D pairs inside given areas (see Fig.~\ref{fig:realnet}). Among them, $F_1$ and $F_2$ are simulated during the first 40min, as $[1, 2, 4, 4, 4, 4, 2, 1]$ unit flows with 5min intervals, while $F_3$ and $F_4$ are generated during a shifted time window from 15min to 55min.

\bt{As the MDP becomes challenging in this experiment, we remove $\wait$ terms in both reward and state, making the value function easier to be fitted. As the price, MARL algorithms will not learn to explicitly optimize the delay. The normalization factor of $\wave$ is 5veh, and that of reward is 20veh per intersection involved in reward calculation.}

\subsubsection{Training results}
\begin{figure}[!h]
  \centering
  \includegraphics[width=.45\textwidth]{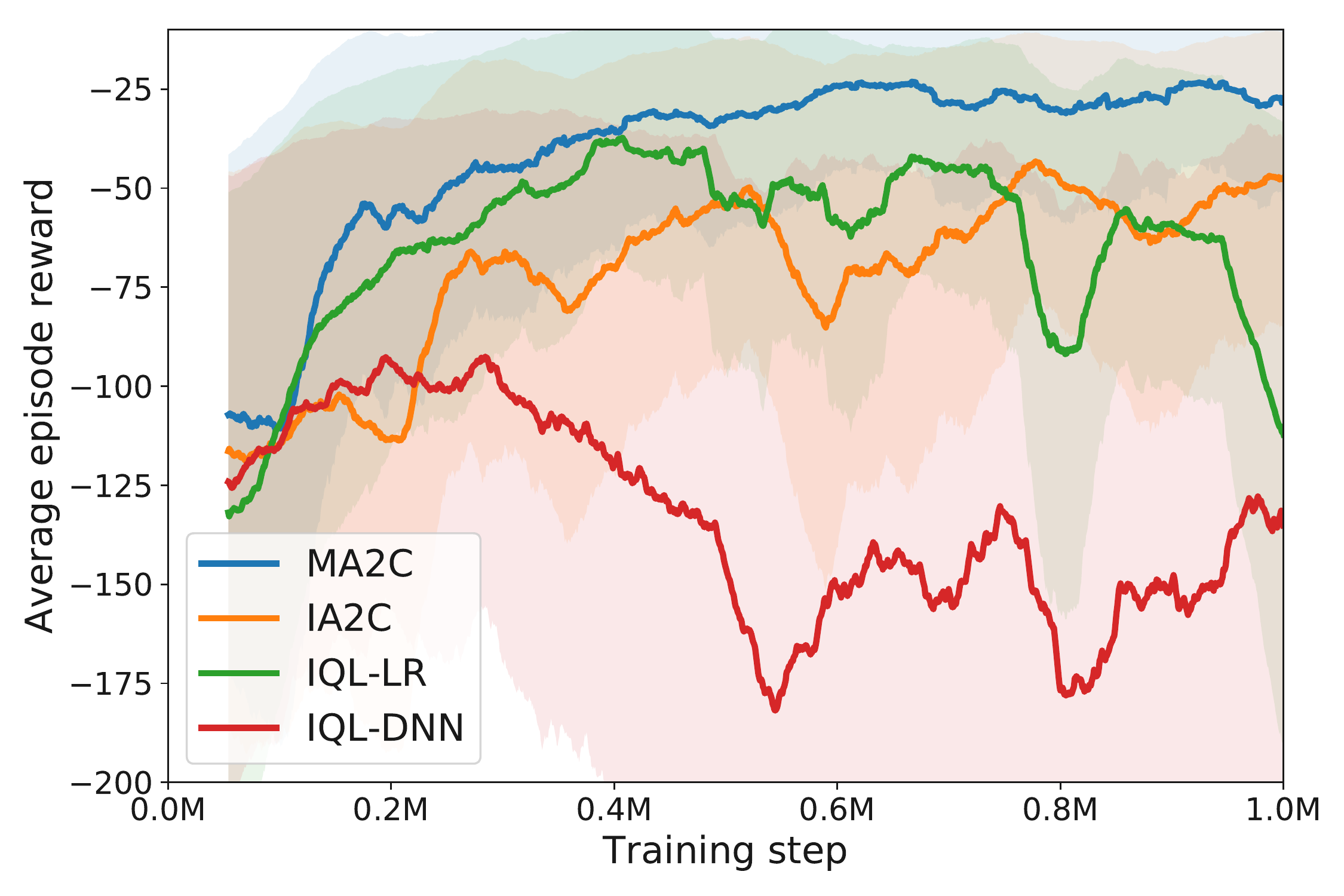}
  \caption{\bt{MARL training curves for Monaco traffic network.}}
  \label{fig:realnet_train}
\end{figure}
\bt{Fig.~\ref{fig:realnet_train} plots the training curves of MARL algorithms. IQL-DNN still does not learn anything, while IQL-LR does not converge, despite good performance in the middle of training. Again, both IA2C and MA2C converge to reasonable policies, and MA2C shows a faster and more stable convergence.}

\subsubsection{Evaluation results}
\bt{Table~\ref{tab:realnet_perf} summarizes the key performance metrics for comparing ATSC in evaluation. The measurements are first spatially aggregated at each time over evaluation episodes, then the temporal average and max are calculated. Except for trip delay, the values are calculated over all evaluated trips. MA2C outperforms other controllers in almost all metrics. Unfortunately, other MARL algorithms are failed to beat the greedy policy in this experiment. Therefore extensive effort and caution is needed before the field deployment of any data-driven ATSC algorithm, regarding its robustness, optimality, efficiency, and safety.}

\begin{figure}[!h]
  \centering
  \includegraphics[width=.45\textwidth]{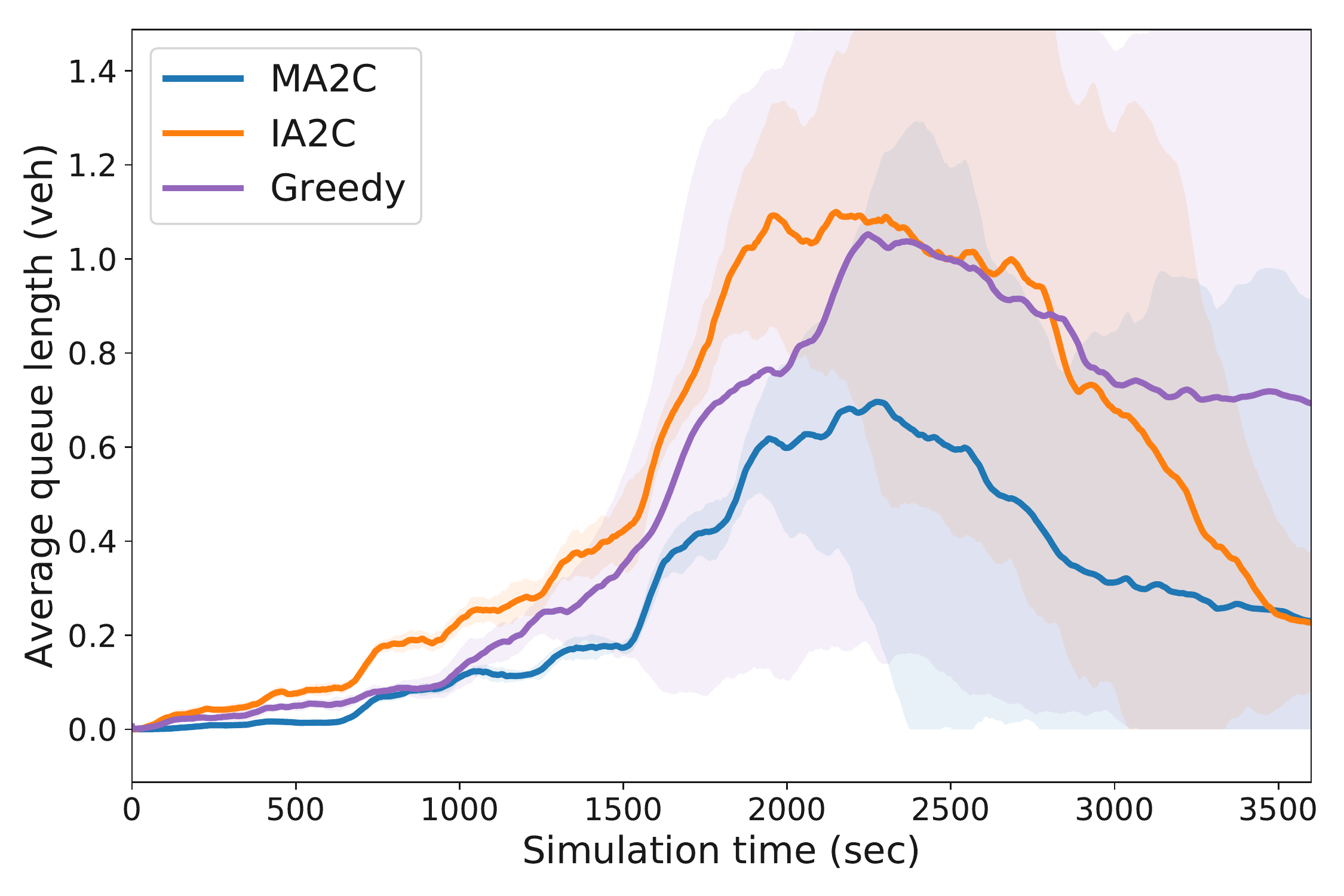}
  \caption{\bt{Average queue length in Monaco traffic network.}}
  \label{fig:real_queue}
\end{figure}

\begin{figure}[!h]
  \centering
  \includegraphics[width=.45\textwidth]{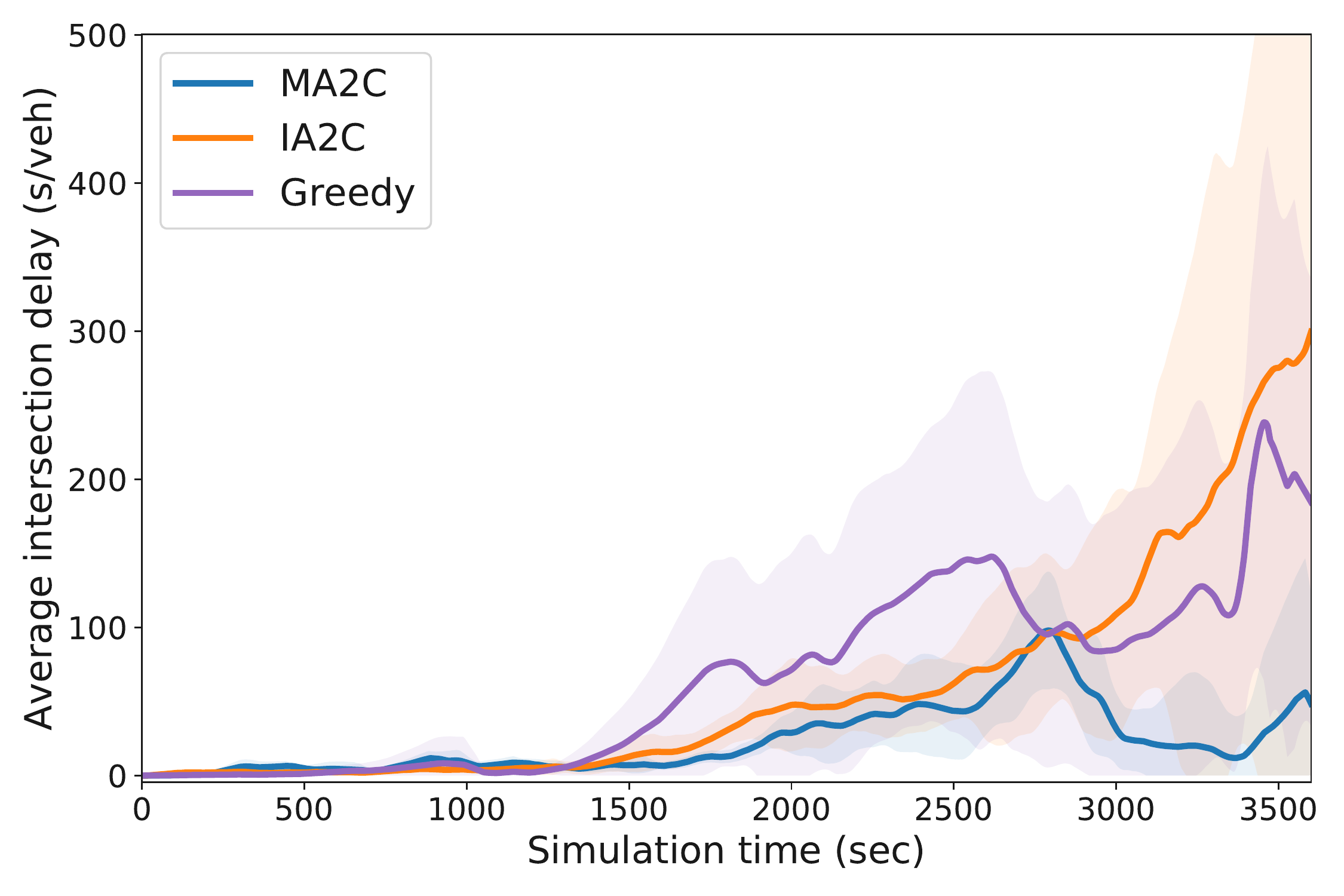}
  \caption{\bt{Average intersection delay in Monaco traffic network.}}
  \label{fig:real_wait}
\end{figure}

IQL algorithms are not included in following analysis as their training performance is as bad as that of random exploration. Fig.~\ref{fig:real_queue} and Fig.~\ref{fig:real_wait} plot the average queue length and average intersection delay over simulation time, under different ATSC policies. \bbt{As expected, both IA2C and Greedy are able to reduce the queue lengths after peak values, and IA2C achieves a better recovery rate. However, both of them are failed to maintain sustainable intersection delays. This may be because of the ``central area'' congestion after upstream intersections greedily maximizing their local flows. On the other hand, MA2C is able to achieve lower and more sustainable intersection delays, by distributing the traffic more homogeneously among intersections via coordination with shared neighborhood fingerprints.}

\begin{figure}[!h]
  \centering
  \includegraphics[width=.45\textwidth]{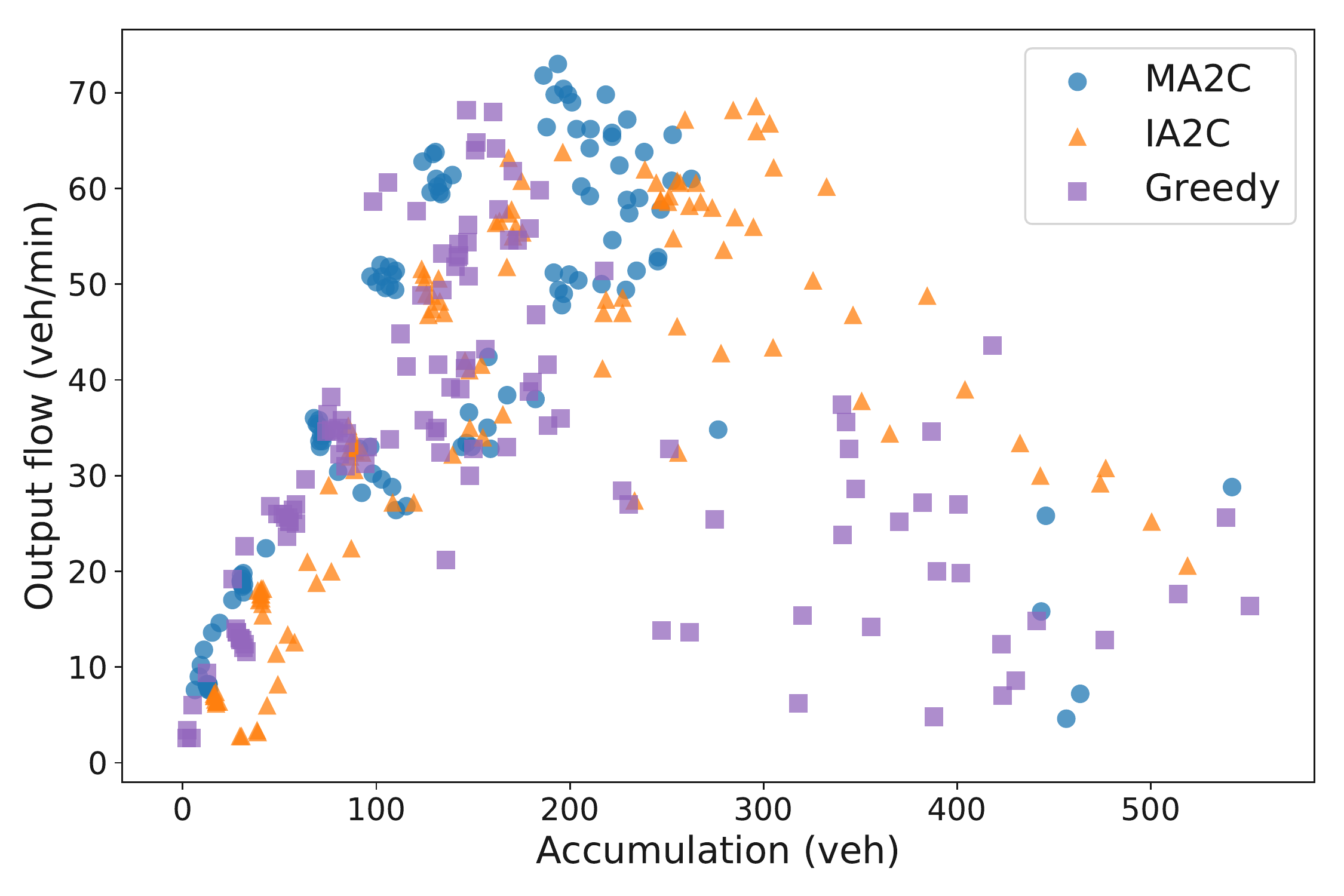}
  \caption{\bt{Output flow vs vehicle accumulation scatter in Monaco traffic network. Each point is aggregated over 5min.}}
  \label{fig:real_mfd}
\end{figure}

\bt{Fig.~\ref{fig:real_mfd} scatters the output (trip completion) flow and network vehicle accumulation for different ATSC algorithms. Macroscopic fundamental diagram (MFD) is present for each controller: when the network density is low, output increases as accumulation grows; when the network becomes more saturated, further accumulation will decrease the output, leading to a potential congestion. As we can see, compared to other controllers, MA2C is able to maintain most points around the ``sweet-spot'', maximizing the utilization of network capacity.}

\section{Conclusions} \label{sec:conclusion}
This paper proposed a novel A2C based MARL algorithm for scalable and robust ATSC. Specifically, 1) fingerprints of neighbors were adapted to improve observability; and 2) a spatial discount factor was introduced to reduce the learning difficulty. \bt{Experiments in a synthetic traffic grid and a Monaco traffic network demonstrated the robustness, optimality, and scalability of the proposed MA2C algorithm, which outperformed other state-of-the-art MARL algorithms.}

\bbt{Non-trivial future works are still remaining for the real-world deployment of proposed MARL algorithm. These include 1)  improving the reality of traffic simulator to provide reliable training data regarding real-world traffic demand and dynamics; 2) improving the algorithm robustness on noisy and delayed state measurements from road sensors; 3) building a pipeline that can train and deploy deep MARL models to each intersection controller for a given traffic scenario; 4) improving the end-to-end pipeline latency, with a focus on the inference time and memory consumption of model query at each intersection, as well as the communication delay among neighboring intersections for state and fingerprint sharing.}

\bibliographystyle{ieeetr}
\bibliography{Ref}

\begin{IEEEbiography}[{\includegraphics[width=25mm,height=32mm,clip,keepaspectratio]{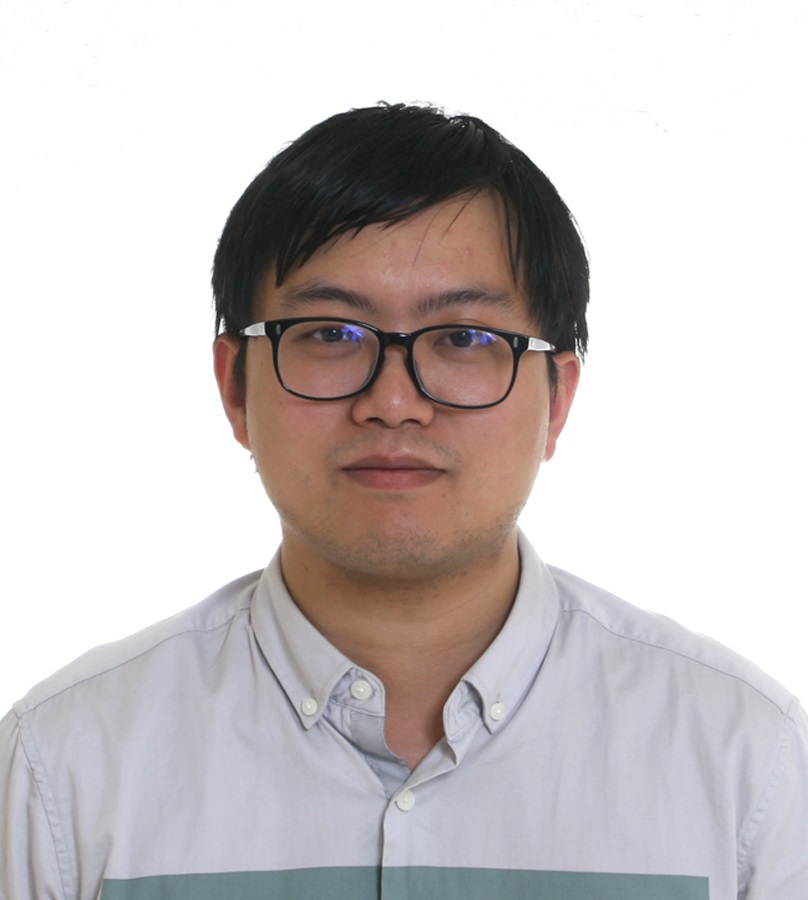}}]{Tianshu Chu}
received the B.S. degree in physics from the Weseda University, Tokyo, Japan, in 2010, and the M.S. and Ph.D degrees from the Department of Civil and Environmental Engineering, Stanford University, in 2012 and 2016. He is currently a data scientist at Uhana Inc. and an adjunct professor at the Stanford Center for Sustainable Development \& Global Competitiveness. His research interests include reinforcement learning, deep learning, multi-agent learning, and their applications to traffic signal control, wireless network control, autonomous driving, and other engineering control systems.
\end{IEEEbiography}

\begin{IEEEbiography}[{\includegraphics[width=25mm,height=32mm,clip,keepaspectratio]{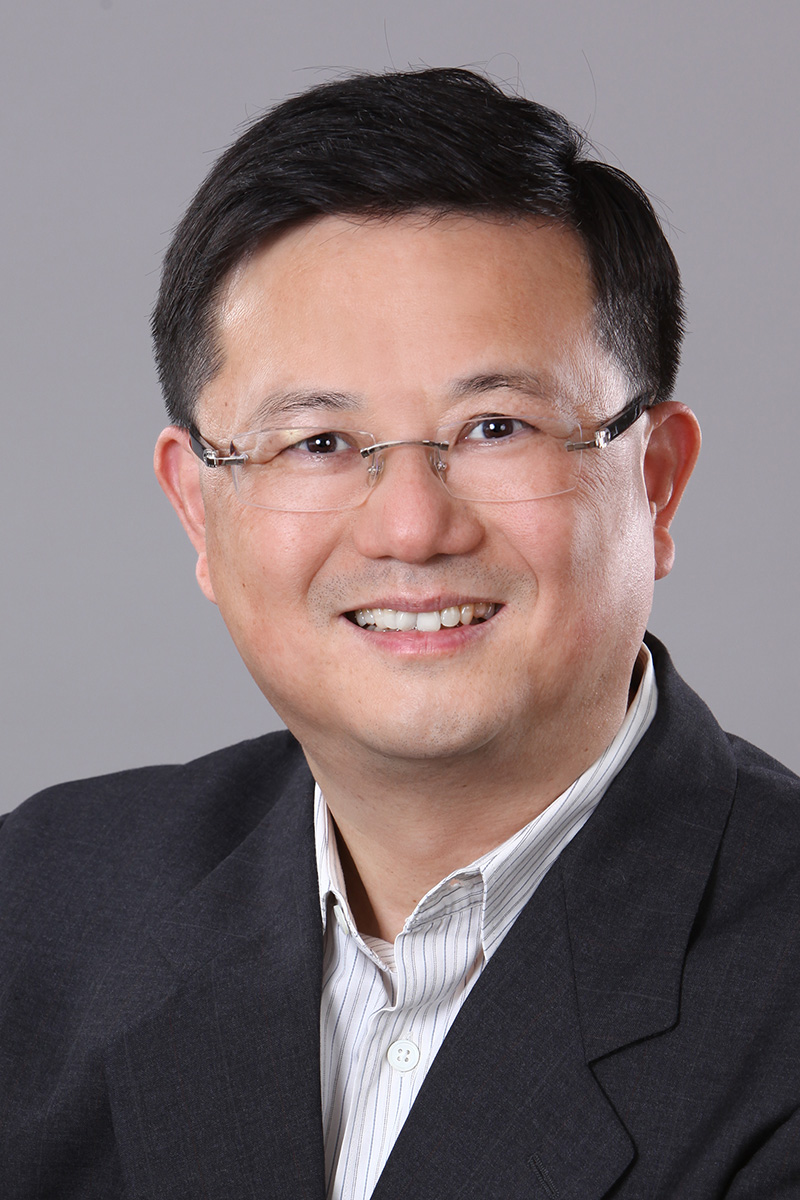}}]{Jie Wang}
received his B.S. degree from Shanghai JiaoTong University, two M.S. degrees from Stanford University and University of Miami, and the Ph.D degree in civil and environmental engineering from Stanford University, in 2003. He is currently an adjunct professor with the Department of Civil and Environmental Engineering, Stanford University, and the executive director of the Stanford Center for Sustainable Development \& Global Competitiveness. His research interests include information and knowledge management for sustainable development and innovation, enterprise IT infrastructure management, smart manufacturing, smart infrastructures and smart city, and environmental informatics.
\end{IEEEbiography}

\begin{IEEEbiography}[{\includegraphics[width=25mm,height=32mm,clip,keepaspectratio]{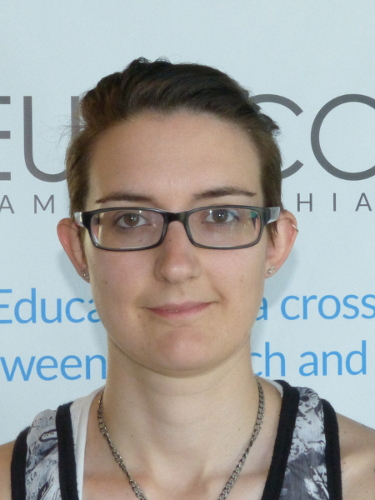}}]{Lara Codec\`a}
received a Ph.D. Degree from the University of Luxembourg in 2016 and her Computer Sciences Master degree at the University of Bologna (Italy) in 2011. In 2011, she was a visiting fellow at Prof. Dr. Mario Gerla's Vehicular Lab at the University of California, Los Angeles (UCLA). She is currently a post-doctoral fellow in the CATS group in EURECOM (France). Her research interests include (Cooperative) Intelligent Transportation Systems, Vehicular Traffic Modelling, and Big-data Analysis. She is active in the SUMO community and collaborates with the SUMO developers at DLR (German Aerospace Centre).
\end{IEEEbiography}

\begin{IEEEbiography}[{\includegraphics[width=25mm,height=32mm,clip,keepaspectratio]{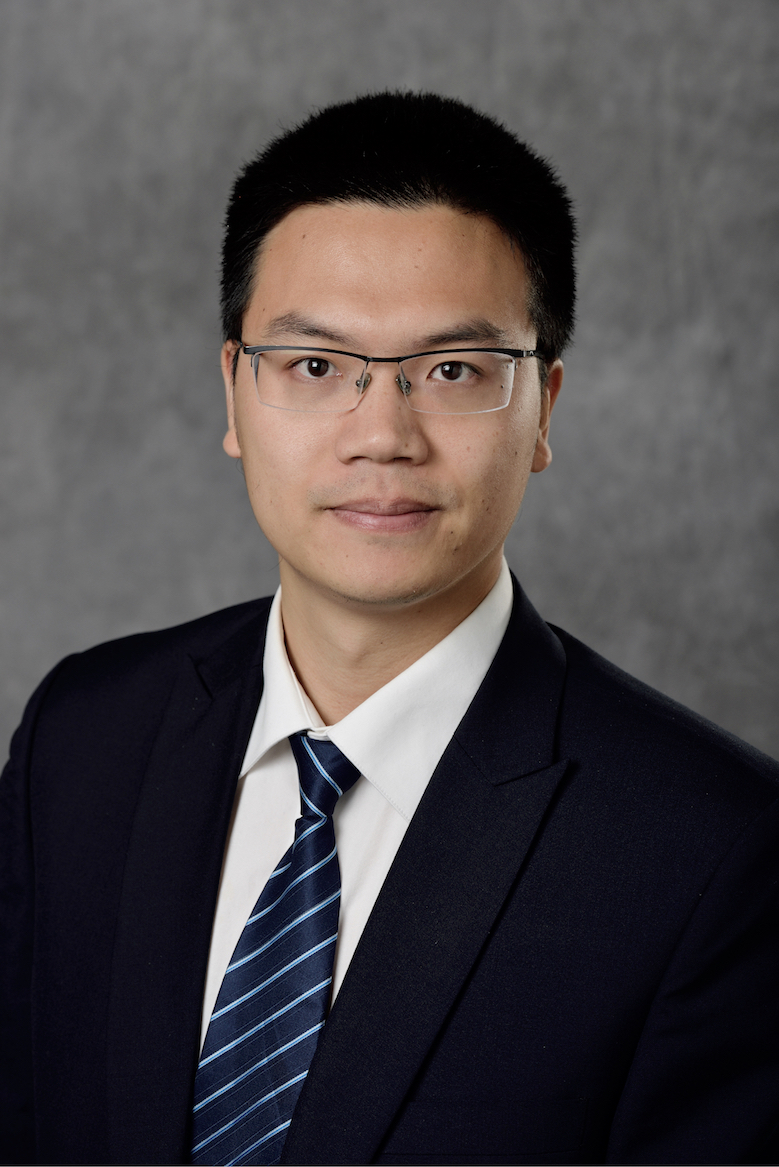}}]{Zhaojian Li}
is an Assistant Professor in the Department of Mechanical Engineering at Michigan State University. He obtained M.S. (2013) and Ph.D. (2015) in Aerospace Engineering (flight dynamics and control) at the University of Michigan, Ann Arbor. As an undergraduate, Dr. Li studied at Nanjing University of Aeronautics and Astronautics, Department of Civil Aviation, in China. Dr. Li worked as an algorithm engineer at General Motors from January 2016 to July 2017. His research interests include Learning-based Control, Nonlinear and Complex Systems, and Robotics and Automated Vehicles. Dr. Li was a recipient of the National Scholarship from China.
\end{IEEEbiography}
\end{document}